\documentstyle[twocolumn]{article}    

\title{A CENTERING APPROACH TO PRONOUNS \\
\begin{small} 25th Annual Meeting of the Association of Computational
Linguistics, 1987 \end{small}}
\author{Susan E. Brennan, Marilyn Walker Friedman, Carl J. Pollard}
\date{Hewlett-Packard Laboratories \\  
      1501 Page Mill Road \\
      Palo Alto, CA 94304, USA}

\begin{document}           
\input{psfig}
\bibliographystyle{alpha}  
\maketitle                 

\begin{abstract}
     In this paper we present a formalization of the centering
approach to modeling attentional structure in discourse and use it as
the basis for an algorithm to track discourse context and bind
pronouns. As described in \cite{GJW86}, the process of centering
attention on entities in the discourse gives rise to the
intersentential transitional states of {\em continuing}, {\em retaining}
and  {\em shifting}.
We propose an extension to these states which handles some
additional cases of multiple ambiguous pronouns.  The algorithm has
been implemented in an HPSG natural language system which serves as
the interface to a database query application.
\end{abstract}

\setlength{\unitlength}{.75in}

\section{Introduction}
\label{intro-sec}

     In the approach to discourse structure developed in \cite{Sidner83a} and
\cite{GJW86}, a discourse exhibits both global and local coherence.  On
this view, a key element of local coherence is {\em centering}, a system of
rules and constraints that govern the relationship between what the
discourse is about and some of the linguistic choices made by the
discourse participants, e.g. choice of grammatical function, syntactic
structure, and type of referring expression (proper noun, definite or
indefinite description, reflexive or personal pronoun, etc.).
Pronominalization in particular serves to focus attention on what is
being talked about; inappropriate use or failure to use pronouns
causes communication to be less fluent.  For instance, it takes longer
for hearers to process a pronominalized noun phrase that is {\em not\/} in
focus than one that is, while it takes longer to process a
non-pronominalized noun phrase that {\em is\/} in focus than one that is not
\cite{Gui85}.

     The \cite{GJW86} centering model is based on the following
assumptions.  A discourse segment consists of a sequence of utterances
$U_{1}, \ldots ,U_{m}$.  With each utterance $U_{n}$ is associated a
list of {\em forward-looking centers}, $Cf(U_{n})$, consisting of
those discourse entities that are {\em directly realized\/} or {\em
realized\/}\footnote {$U$ {\em directly realizes} {\bf c} if U is an
utterance (of some phrase, not necessarily a full clause) for which
{\bf c} is the semantic interpretation, and $U$ {\em realizes\/} {\bf
c} if either {\bf c} is an element of the situation described by the
utterance $U$ or {\bf c} is directly realized by some subpart of $U$.
Realizes is thus a generalization of directly realizes\cite{GJW86}.}
by linguistic expressions in the utterance. Ranking of an entity on
this list corresponds roughly to the likelihood that it will be the
primary focus of subsequent discourse; the first entity on this list
is the {\em preferred center}, $Cp(U_{n})$.  $U_{n}$ actually centers,
or is ``about'', only one entity at a time, the {\em backward-looking
center}, $Cb(U_{n})$.  The backward center is a confirmation of an
entity that has already been introduced into the discourse; more
specifically, it must be realized in the immediately preceding
utterance, $U_{n-1}$. There are several distinct types of transitions
from one utterance to the next. The typology of transitions is based
on two factors: whether or not the center of attention, $Cb$, is the
same from $U_{n-1}$ to $U_{n}$, and whether or not this entity
coincides with the preferred center of $U_{n}$. Definitions of
these transition types appear in figure \ref{state-fig}.

\begin{figure*}
\begin{center}

\begin{picture}(4,2)

\put(0,0){\framebox(2,1){RETAINING}}
\put(0,1){\framebox(2,1){CONTINUING}}
\put(2,0){\framebox(2,2){SHIFTING}}

\put(1,2.2){\makebox(0,0){$Cb(U_{n}) = Cb(U_{n-1})$}}
\put(3,2.2){\makebox(0,0){$Cb(U_{n}) \neq Cb(U_{n-1})$}}
\put(-.9,1.5){\makebox(0,0){$Cb(U_{n}) = Cp(U_{n})$}}
\put(-.9,.5){\makebox(0,0){$Cb(U_{n}) \neq Cp(U_{n})$}}

\end{picture}
\end{center}

\caption{Transition States}
\label{state-fig}
\end{figure*}

     These transitions describe how utterances are linked together in
a coherent local segment of discourse.  If a speaker has a number of
propositions to express, one very simple way to do this coherently is
to express all the propositions about a given entity ({\em
continuing}) before introducing a related entity ({\em retaining}) and
then {\em shifting} the center to this new entity. See figure
\ref{alg-ex-fig}.  Retaining may be a way to signal an intention to
shift.  While we do not claim that speakers really behave in such an
orderly fashion, an algorithm that expects this kind of behavior is
more successful than those which depend solely on recency or
parallelism of grammatical function.  The interaction of centering
with global focusing mechanisms and with other factors such as
intentional structure, semantic selectional restrictions, verb tense
and aspect, modality, intonation and pitch accent are topics for
further research.

Note that these transitions are more specific than focus movement
as described in \cite{Sidner83a}.  The extension we propose makes
them more specific still.  Note also that the $Cb$ of \cite{GJW86}
corresponds roughly to Sidner's discourse focus and the $Cf$ to her
potential foci.

     The formal system of constraints and rules for centering, as we
have interpreted them from \cite{GJW86}, are as follows. For each
$U_{n}$ in $U_{1}, \ldots ,U_{m}$:
\begin{itemize}
\item
{\bf CONSTRAINTS}
   \begin{enumerate}
      \item There is precisely one $Cb$.
      \item Every element of $Cf(U_{n})$ must be realized in $U_{n}$.
      \item $Cb(U_{n})$ is the highest-ranked element of $Cf(U_{n-1})$ that
                 is realized in $U_{n}$.
    \end{enumerate}
\end{itemize}

\newpage
\begin{itemize}
\item
{\bf RULES}
   \begin{enumerate}
     \item If some element of $Cf(U_{n-1})$ is realized as a pronoun in
$U_{n}$,
              then so is $Cb(U_{n})$.
     \item {\em Continuing\/} is preferred over {\em retaining\/} which is
preferred over
             {\em shifting}.
   \end{enumerate}
\end{itemize}

     As is evident in constraint 3, ranking of the items on the
forward center list, $Cf$, is crucial.  We rank the items in $Cf$ by
obliqueness of grammatical relation of the subcategorized
functions of the main verb: that is, first the subject, object, and
object2, followed by other subcategorized functions, and finally,
adjuncts.  This captures the idea in \cite{GJW86} that subjecthood
contributes strongly to the priority of an item on the $Cf$ list.

\begin{figure}[htb]
CONTINUING... \\
{\raggedright
\begin{tabular}{rp{66mm}}
$U_{n+1}$: & Carl works at HP on the Natural Language Project. \\
$     Cb$: & [POLLARD:Carl]  \\
$     Cf$: & ([POLLARD:Carl] [HP:HP] \\
           &  \ [NATLANG:Natural Language Project])
\end{tabular}
}

CONTINUING... \\
\begin{tabular}{rp{66mm}}
$U_{n+2}$: & He manages Lyn. \\
$     Cb$: & [POLLARD:Carl] \\
$     Cf$: & ([POLLARD:A1] [FRIEDMAN:Lyn])
\end{tabular}

He = Carl

CONTINUING... \\
\begin{tabular}{rp{66mm}}
$U_{n+3}$: & He promised to get her a raise. \\
$     Cb$: & [POLLARD:A1] \\
$     Cf$: & ([POLLARD:A2] [FRIEDMAN:A3] \\
           & \  [RAISE:X1])
\end{tabular}

He = Carl, her = Lyn

RETAINING... \\
\begin{tabular}{rp{66mm}}
$U_{n+4}$: & She doesn't believe him. \\
$     Cb$: & [POLLARD:A2] \\
$     Cf$: & ([FRIEDMAN:A4] [POLLARD:A5])
\end{tabular}

She = Lyn, him = Carl

\caption{}
\label{alg-ex-fig}
\end{figure}

     We are aware that this ranking usually coincides with surface
constituent order in English.  It would be of interest to examine data
from languages with relatively freer constituent order (e.g. German)
to determine the influence of constituent order upon centering when
the grammatical functions are held constant.  In addition, languages
that provide an identifiable topic function (e.g. Japanese) suggest
that topic takes precedence over subject.

\begin{figure*}
\begin{center}

\begin{picture}(4,2)

\put(0,0){\framebox(2,1){RETAINING}}
\put(0,1){\framebox(2,1){CONTINUING}}
\put(2,1){\framebox(2,1){SHIFTING-1}}
\put(2,0){\framebox(2,1){SHIFTING}}

\put(1,2.2){\makebox(0,0){Cb(Un) = Cb(Un-1)}}
\put(3,2.2){\makebox(0,0){Cb(Un) $\neq$ Cb(Un-1)}}
\put(-.9,1.5){\makebox(0,0){Cb(Un) = Cp(Un)}}
\put(-.9,.5){\makebox(0,0){Cb(Un) $\neq$ Cp(Un)}}

\end{picture}
\end{center}

\caption{Extended Transition States}
\label{state-fig2}
\end{figure*}

     The part of the HPSG system that uses the centering algorithm for
pronoun binding is called the pragmatics processor. It interacts with
another module called the semantics processor, which computes
representations of intrasentential anaphoric relations, (among other
things).  The semantics processor has access to information such as
the surface syntactic structure of the utterance.  It provides the
pragmatics processor with representations which include of a set of
reference markers. Each reference marker is contraindexed\footnote{See
\cite{BP80} and \cite{CHOM80} for conditions on coreference.} with
expressions with which it cannot co-specify\footnote{See \cite{Sidner83a}
for definition and discussion of co-specification.  Note that this use
of co-specification is not the same as that used in \cite{Sel85}}.
Reference markers also carry information about agreement and
grammatical function. Each pronominal reference marker has a unique
index from $A_{1}, \ldots ,A_{n}$ and is displayed in the figures in
the form [POLLARD:A1], where POLLARD is the semantic representation of
the co-specifier.  For non-pronominal reference markers the surface
string is used as the index. Indices for indefinites are generated
from $X_{1}, \ldots ,X_{n}$.

\section{ Extension}
\label{extn-sec}

     The constraints proposed by \cite{GJW86} fail in certain
examples like the following (read with pronouns destressed):

\begin{verse}
   Brennan drives an Alfa Romeo.\\
   She drives too fast.\\
   Friedman races her on weekends.\\
   She often beats her.
\end{verse}

     This example is characterized by its multiple ambiguous pronouns
and by the fact that the final utterance achieves a shift (see figure
\ref{extn-fig}).  A shift is inevitable because of constraint 3, which
states that the $Cb(U_{n})$ must equal the $Cp(U_{n-1})$ (since the
$Cp(U_{n-1})$ is directly realized by the subject of $U_{n}$,
``Friedman'').  However the constraints and rules from \cite{GJW86}
would fail to make a choice here between the co-specification
possibilities for the pronouns in $U_{n}$.  Given that the transition
is a shift, {\em there seem to be more and less coherent ways to
shift}. Note that the three items being examined in order to
characterize the transition between each pair of {\em
anchors\/}\footnote {An anchor is a $<Cb, Cf>$ pair for an utterance}
are the $Cb$ of $U_{n-1}$, the $Cb$ of $U_{n}$, and the $Cp$ of
$U_{n}$.  By \cite{GJW86} a shift occurs whenever successive $Cb$'s
are not the same.  This definition of shifting does {\em not} consider
whether the $Cb$ of $U_{n}$ and the $Cp$ of $U_{n}$ are equal.  It
seems that the status of the $Cp$ of $U_{n}$ should be as important in
this case as it is in determining the retaining/chontinuing
distinction.

     Therefore, we propose the following extension which handles some
additional cases containing multiple ambiguous pronouns: we have
extended rule 2 so that there are two kinds of shifts. A transition
for $U_{n}$ is ranked more highly if $Cb(U_{n}) = Cp(U_{n})$; this
state we call {\em shifting-1} and it represents a more coherent way to
shift.  The preferred ranking is $continuing \succ retaining \succ \
${\em shifting-1}$ \  \succ shifting$ (see figure \ref{state-fig2}).
This extension enables us to successfully bind the ``she'' in the
final utterance of the example in figure \ref{extn-fig} to
``Friedman.'' The appendix illustrates the application of the algorithm
to figure \ref{extn-fig}.

\begin{figure}[tb]

CONTINUING... \\
\begin{tabular}{rp{66mm}}
$U_{n+1}$: & Brennan drives an Alfa Romeo. \\
$     Cb$: & [BRENNAN:Brennan]  \\
$     Cf$: & ([BRENNAN:Brennan] [X2:Alfa Romeo])
\end{tabular}

CONTINUING... \\
\begin{tabular}{rp{66mm}}
$U_{n+2}$: & She drives too fast. \\
$     Cb$: & [BRENNAN:Brennan]  \\
$     Cf$: & ([BRENNAN:A7])
\end{tabular}

She = Brennan

RETAINING... \\
\begin{tabular}{rp{66mm}}
$U_{n+3}$: & Friedman races her on weekends. \\
$     Cb$: & [BRENNAN:A7]   \\
$     Cf$: & ([FRIEDMAN:Friedman] [BRENNAN:A8] \\
           & \ [WEEKEND:X3])
\end{tabular}

her = Brennan

SHIFTING-1... \\
\begin{tabular}{rp{66mm}}
$U_{n+4}$: & She often beats her. \\
$     Cb$: & [FRIEDMAN:Friedman] \\
$     Cf$: & ([FRIEDMAN:A9] [BRENNAN:A10])
\end{tabular}

She = Friedman, her = Brennan

\caption{}
\label{extn-fig}
\end{figure}

     Kameyama \cite{Kameyama86b} has proposed another extension to the
\cite{GJW86} theory -- a property-sharing constraint which attempts to
enforce a parallellism between entities in successive utterances.  She
considers two properties: {\em SUBJ\/} and {\em IDENT\/}.  With her
extension, subject pronouns prefer subject antecedents and non-subject
pronouns prefer non-subject antecedents. However, structural
parallelism is a consequence of our ordering the $Cf$ list by
grammatical function and the preference for continuing over retaining.
Furthermore, the constraints suggested in \cite{GJW86} succeed in many
cases {\em without\/} invoking an independent structural parallelism
constraint, due to the distinction between continuing and retaining,
which Kameyama fails to consider.  Her example which we reproduce in
figure \ref{kam-figa} can also be accounted for using the
continuing/retaining distinction\footnote {It seems that property
sharing of IDENT is still necessary to account for logophoric use of
pronouns in Japanese.}. The third utterance in this example has two
interpretations which are {\em both} consistent with the centering
rules and constraints.  Because of rule 2, the interpretation in
figure \ref{kam-figa} is preferred over the one in figure
\ref{kam-figb}.

\begin{figure}[tb]

CONTINUING... \\
\begin{tabular}{rp{66mm}}
$U_{n+1}$: & Who is Max waiting for? \\
$     Cb$: & [PLANCK:Max] \\
$     Cf$: & ([PLANCK:Max])
\end{tabular}

CONTINUING... \\
\begin{tabular}{rp{66mm}}
$U_{n+2}$: & He is waiting for Fred. \\
$     Cb$: & [PLANCK:Max]   \\
$     Cf$: & ([PLANCK:A1] [FLINTSTONE:Fred])
\end{tabular}

He = Max
CONTINUING... \\
\begin{tabular}{rp{66mm}}
$U_{n+3}$: & He invited him to dinner. \\
$     Cb$: & [PLANCK:A1] \\
$     Cf$: & ([PLANCK:A2] [FLINTSTONE:A3])
\end{tabular}

He = Max, him = Fred

\caption{}
\label{kam-figa}
\end{figure}

\section{Algorithm for centering and pronoun binding}
\label{alg-sec}

\begin{figure}[tb]

CONTINUING... \\
\begin{tabular}{rp{66mm}}
$U_{n+1}$: & Who is Max waiting for? \\
$     Cb$: & [PLANCK:Max]  \\
$     Cf$: & ([PLANCK:Max])
\end{tabular}

CONTINUING... \\
\begin{tabular}{rp{66mm}}
$U_{n+2}$: & He is waiting for Fred. \\
$     Cb$: & [PLANCK:Max]  \\
$     Cf$: & ([PLANCK:A1] [FLINTSTONE:Fred])
\end{tabular}

he = Max

RETAINING... \\
\begin{tabular}{rp{66mm}}
$U_{n+3}$: & He invited him to dinner. \\
$     Cb$: & [PLANCK:A1]  \\
$     Cf$: & ([FLINTSTONE:A3] [PLANCK:A2])
\end{tabular}

He = Fred, him = Max

\caption{}
\label{kam-figb}
\end{figure}

     There are three basic phases to this algorithm.  First the
proposed anchors are {\em constructed}, then they are {\em filtered},
and finally, they are {\em classified} and {\em ranked}. The proposed
anchors represent all the co-specification relationships available for
this utterance.

\begin{figure*}
\caption{Algorithm and Example}
\label{alg-fig}
\end{figure*}

     Each step is discussed and illustrated in figure \ref{alg-fig}.
It would be possible to classify and rank the proposed anchors before
filtering them without any other changes to the algorithm.  In fact,
using this strategy one could see if the highest ranked proposal
passed all the filters, or if the next highest did, etc.  The three
filters in the filtering phase may be done in parallel. The example we
use to illustrate the algorithm is in figure \ref{alg-ex-fig}.

\section{Discussion}
\label{disc-sec}

\subsection{Discussion of the algorithm}

     The goal of the current algorithm design was conceptual clarity
rather than efficiency.  The hope is that the structure provided will
allow easy addition of further constraints and preferences.  It would
be simple to change the control structure of the algorithm so that it
first proposed all the continuing or retaining anchors and then the
shifting ones, thus avoiding a precomputation of all possible anchors.

     \cite{GJW86} states that a realization may contribute more than
one entity to the $Cf(U)$. This is true in cases when a partially
specified semantic description is consistent with more than one
interpretation.  There is no need to enumerate explicitly all the
possible interpretations when constructing possible
$Cf(U)$'s\footnote{Barbara Grosz, personal communication, and
\cite{GJW86}}, as long as the associated semantic theory allows
partially specified interpretations. This also holds for entities not
directly realized in an utterance.  On our view, after referring to
``a house'' in $U_{n}$, a reference to ``the door'' in $U_{n+1}$ might
be gotten via inference from the representation for ``a house'' in
$Cf(U_{n})$. Thus when the proposed anchors are constructed there is
no possibility of having an infinite number of potential $Cf$'s for an
utterance of finite length.

     Another question is whether the preference ordering of
transitions in constraint 3 should always be the same.  For some
examples, particularly where $U_{n}$ contains a single pronoun and
$U_{n-1}$ is a retention, some informants seem to have a preference
for shifting, whereas the centering algorithm chooses a continuation
(see figure \ref{anom-fig}).  Many of our informants have no strong
preference as to the co-specification of the unstressed ``She'' in
$U_{n+4}$.  Speakers can avoid ambiguity by stressing a pronoun
with respect to its phonological environment.  A computational system
for {\em understanding} may need to explicitly acknowledge this
ambiguity.

\begin{figure}[htb]

CONTINUING... \\
\begin{tabular}{rp{66mm}}
$U_{n+1}$: & Brennan drives an Alfa Romeo. \\
$     Cb$: & [BRENNAN:Brennan]  \\
$     Cf$: & ([BRENNAN:Brennan] [ALFA:X1])
\end{tabular}

CONTINUING... \\
\begin{tabular}{rp{66mm}}
$U_{n+2}$: & She drives too fast. \\
$     Cb$: & [BRENNAN:Brennan] \\
$     Cf$: & ([BRENNAN:A7])
\end{tabular}

She = Brennan

RETAINING... \\
\begin{tabular}{rp{66mm}}
$U_{n+3}$: & Friedman races her on weekends. \\
$     Cb$: & [BRENNAN:A7] \\
$     Cf$: & ([FRIEDMAN:Friedman] [BRENNAN:A8]) \\
           & \ [WEEKEND:X3])
\end{tabular}

her = Brennan

CONTINUING... \\
\begin{tabular}{rp{66mm}}
$U_{n+4}$: & She goes to Laguna Seca. \\
$     Cb$: & [BRENNAN:A8]  \\
$     Cf$: & ([BRENNAN:A9] \\
           & [LAG-SEC:Laguna Seca])
\end{tabular}

She = Brennan??

\caption{}
\label{anom-fig}
\end{figure}

     A computational system for {\em generation} would try to plan a
retention as a signal of an impending shift, so that after a
retention, a shift would be preferred rather than a continuation.

\subsection{Future Research}

     Of course the local approach described here does not provide all
the necessary information for interpreting pronouns; constraints are
also imposed by world knowledge, pragmatics, semantics and phonology.

     There are other interesting questions concerning the centering
algorithm. How should the centering algorithm interact with an
inferencing mechanism?  Should it make choices when there is more than
one proposed anchor with the same ranking?  In a database query
system, how should answers be incorporated into the discourse model?
How does centering interact with a treatment of definite/indefinite
NP's and quantifiers?

     We are exploring ideas for these and other extensions to the
centering approach for modeling reference in local discourse.

\section{Acknowledgements}

     We would like to thank the following people for their help and
insight: Hewlett Packard Lab's Natural Language group, CSLI's DIA
group, Candy Sidner, Dan Flickinger, Mark Gawron, John Nerbonne, Tom
Wasow, Barry Arons, Martha Pollack, Aravind Joshi, two anonymous referees, and
especially Barbara Grosz.

\section{Appendix}
 This illustrates the extension in the same detail as the example
 we used in the algorithm. The numbering here corresponds to the
 numbered steps in the algorithm figure \ref{alg-fig}. The example
 is the last utterance from figure \ref{extn-fig}.

{\bf EXAMPLE:} She often beats her.

\begin{enumerate}

\item {\bf  CONSTRUCT THE PROPOSED ANCHORS}
  \begin{enumerate}
  \item ([A9] [A10])
  \item ([A9] [A10])
  \item (([FRIEDMAN:A9] [FRIEDMAN:A10]) \\
          \ ([FRIEDMAN:A9] [BRENNAN:A10]) \\
          \ ([BRENNAN:A9]  [BRENNAN:A10]) \\
          \ ([BRENNAN:A9]  [FRIEDMAN:A10]))

  \item ([FRIEDMAN:Friedman] [BRENNAN:A8]\\
			     \ [WEEKEND:X3] NIL)
  \item  There are 16 possible $<Cb, Cf>$ pairs for this utterance.
      \begin{enumerate}

       \item $<${\small  [FRIEDMAN:Friedman],\\
			 \ \ \ \ ([FRIEDMAN:A9] [FRIEDMAN:A10])}$>$
       \item $<${\small  [FRIEDMAN:Friedman], \\
			 \ \ \ \ ([FRIEDMAN:A9] [BRENNAN:A10])}$>$
       \item $<${\small  [FRIEDMAN:Friedman],\\
			 \ \ \ \ ([BRENNAN:A9] [FRIEDMAN:A10])}$>$
       \item $<${\small  [FRIEDMAN:Friedman], \\
			 \ \ \ \ ([BRENNAN:A9] [BRENNAN:A10])}$>$
       \item $<${\small  [BRENNAN:A8],\\
			 \ \ \ \ ([FRIEDMAN:A9] [FRIEDMAN:A10])}$>$
       \item $<${\small  [BRENNAN:A8], \\
			 \ \ \ \ ([FRIEDMAN:A9] [BRENNAN:A10])}$>$
       \item $<${\small  [BRENNAN:A8],\\
			 \ \ \ \ ([BRENNAN:A9] [FRIEDMAN:A10])}$>$
       \item $<${\small  [BRENNAN:A8],\\
			 \ \ \ \ ([BRENNAN:A9] [BRENNAN:A10])}$>$
       \item $<${\small  [WEEKEND:X3],\\
			 \ \ \ \ ([FRIEDMAN:A9] [FRIEDMAN:A10])}$>$
       \item $<${\small  [WEEKEND:X3],\\
			 \ \ \ \ ([FRIEDMAN:A9] [BRENNAN:A10])}$>$
       \item $<${\small  [WEEKEND:X3],\\
			 \ \ \ \ ([BRENNAN:A9] [FRIEDMAN:A10])}$>$
       \item $<${\small  [WEEKEND:X3],\\
			 \ \ \ \ ([BRENNAN:A9] [BRENNAN:A10])}$>$
       \item $<${\small  NIL,\\
			 \ \ \ \ ([FRIEDMAN:A9] [FRIEDMAN:A10])}$>$
       \item $<${\small  NIL,\\
			 \ \ \ \ ([FRIEDMAN:A9] [BRENNAN:A10])}$>$
       \item $<${\small  NIL,\\
			 \ \ \ \ ([BRENNAN:A9] [FRIEDMAN:A10])}$>$
       \item $<${\small  NIL,\\
			 \ \ \ \ ([BRENNAN:A9] [BRENNAN:A10])}$>$
       \end{enumerate}

  \end{enumerate}

\item {\bf  FILTER THE PROPOSED ANCHORS}

   \begin{enumerate}
   \item  Filter by contraindices.
    Anchors {\em i}, {\em iv}, {\em v}, {\em viii}, {\em ix}, {\em xii}, {\em
xiii}, {\em xvi}
    are eliminated since [A9] and [A10] are contraindexed.

   \item Constraint 3 filter eliminates proposed anchors {\em vii}, {\em ix}
through {\em xvi}.

   \item Rule 1 filter eliminates proposed anchors {\em ix} through {\em xvi}.

   \end{enumerate}

\item {\bf CLASSIFY and RANK}
   \begin{enumerate}
   \item After filtering there are only two anchors left.

\begin{tabular}{rp{66mm}}
{\em ii}:  &  $<${\small  [FRIEDMAN:Friedman],} \\
           & \ {\small ([FRIEDMAN:A9] [BRENNAN:A10])}$>$ \\
{\em iii}: & $<${\small  [FRIEDMAN:Friedman],} \\
           & \ {\small ([BRENNAN:A9] [FRIEDMAN:A10])}$>$
\end{tabular}

     Anchor {\em ii} is classified as shifting-1 whereas  anchor {\em iii} is
     classified as shifting.
   \item Anchor {\em ii} is more highly ranked.
   \end{enumerate}

\end {enumerate}

\nocite{GS86}
\nocite{GJW83}
\nocite{SH84}
\nocite{Sidner81}
\nocite{JW81}

\end{document}